\documentclass{article}

\usepackage{arxiv}

\usepackage[utf8]{inputenc} % allow utf-8 input
\usepackage[T1]{fontenc}    % use 8-bit T1 fonts
\usepackage{hyperref}       % hyperlinks
\usepackage{url}            % simple URL typesetting
\usepackage{booktabs}       % professional-quality tables
\usepackage{amsfonts}       % blackboard math symbols
\usepackage{nicefrac}       % compact symbols for 1/2, etc.
\usepackage{microtype}      % microtypography
\usepackage{lipsum}
\usepackage{graphicx}
\usepackage{amsmath}
\title{Multimodal Prompt Learning with Irregular EHRs for Robust Monitoring of Critical Care Patients}
\author{Yixin Yang\textsuperscript{1,$*$} \quad Yueyang Sun\textsuperscript{2,}\thanks{Equal Contribution} \quad Weichen Liu\textsuperscript{3} \quad Xianbing Zhao\textsuperscript{4} \quad Sicen Liu\textsuperscript{1,}\thanks{Corresponding Author}\\
\textsuperscript{1}Faculty of Engineering, Shenzhen MSU-BIT University, \\
\textsuperscript{2}International School, Beijing University of Posts and Telecommunications,\\
\textsuperscript{3}School of Computer Science and Engineering, Southeast University,\\
\textsuperscript{4}School of Artificial Intelligence and Computer Science, Jiangnan University\\
  \tt\small{yyx@smbu.edu.cn, 2474931434@bupt.edu.cn, 213233033@seu.edu.cn,} \\ \tt\small{zhaoxianbing\_hitsz@163.com, liusc@smbu.edu.cn}
}

\begin{document}

\maketitle

\begin{abstract}
Accurate assessment of patients in intensive care units (ICUs) is essential for timely clinical intervention and improved patient outcomes. Multimodal electronic health records (EHRs), including structured physiological time series and longitudinal clinical notes, provide complementary information for critical care prediction. However, in real-world clinical settings, individual modalities may be partially observed or entirely unavailable, resulting in substantial performance degradation for existing multimodal models. To address this challenge, we propose a multimodal prompt-learning framework for robust clinical prediction under diverse missing-modality scenarios. The proposed framework introduces four complementary types of prompts: generative prompts, missing-signal prompts, missing-type prompts, and temporal prompts. Generative prompts construct surrogate latent representations for unavailable modalities, while missing-signal prompts distinguish observed representations from generated ones. Missing-type prompts condition the model on different modality-availability configurations, whereas temporal prompts perform condition-specific aggregation over temporally encoded clinical sequences. Together, these prompts enable the model to capture missingness-aware intramodal dependencies and cross-modal interactions within a unified architecture. Extensive experiments demonstrate that our method outperforms existing approaches across evaluation metrics on two missingness settings. Ablation and robustness analyses further verify the complementary contributions of the four prompt types and the effectiveness of the proposed framework for clinical prediction from incomplete multimodal EHR data.
 
\end{abstract}

%%%%%%%%%%%%%%%%%%%%%%%%%%%%%%%%%%%%%%%%%%%%%%%%%%%%%%%%%

\section{Introduction}
Intensive care units (ICUs) provide continuous care for patients with life-threatening conditions, such as severe trauma~\cite{tisherman2018icu}, sepsis~\cite{alberti2002epidemiology}, and organ failure~\cite{afessa2007severity}. The early hours following ICU admission are particularly critical, as patients may experience rapid physiological deterioration and are highly vulnerable to delayed or inappropriate clinical decisions~\cite{cullen1997preventable,otero2006preventable}. Accurate prediction of in-hospital mortality based on patients' clinical information collected during the first 48 hours of ICU admission is therefore essential for timely risk assessment, resource allocation, and clinical intervention.
The widespread adoption of electronic health records (EHRs) has substantially transformed the documentation and analysis of patients' health conditions in ICUs~\cite{adler2015electronic,atasoy2019digitization,rajkomar2018scalable}. EHRs contain heterogeneous and complementary sources of clinical information, including multivariate irregularly sampled time series (MISTS), such as laboratory measurements and vital signs, as well as longitudinal clinical notes documenting patients' symptoms, diagnoses, treatments, and disease progression. The increasing availability of these data has facilitated the development of deep learning models for various clinical applications~\cite{xiao2018opportunities,ahmed2023deep,shickel2017deep}, including mortality prediction~\cite{xie2020autoscore}, sepsis onset prediction~\cite{valik2023predicting},phenotype classification~\cite{hourani2023label}, and medication recommendation~\cite{zhang2023knowledge,liu2023shape}.

\begin{figure}[t]
  \centering
  \includegraphics[width=0.8\linewidth]{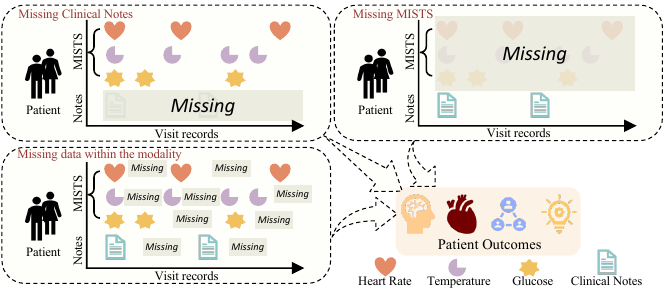}
  \caption{Examples of different types of missing information in EHR data include missing clinical notes, missing MISTS, and missing data within the modality. The heart rate and temperature are monitored regularly with different frequencies, and glucose is a laboratory test ordered at irregular time intervals. The clinical notes are free text.}
  \label{fig:missing_case}
\end{figure}

Existing studies have developed multimodal architectures to integrate irregular physiological measurements and clinical notes~\cite{zhang2023improving,liu2024knowledge}. However, both modalities may not always be available during inference, as real-world EHRs may contain incomplete data or have one modality entirely missing due to equipment failures, delayed documentation, data corruption, privacy restrictions, or variations in clinical workflows. As illustrated in Figure~\ref{fig:missing_case}, different missing-modality configurations introduce distinct information gaps and may substantially degrade mortality prediction. Moreover, observations within each modality are temporally irregular: physiological variables are recorded at uneven intervals, while clinical notes are documented at different stages of an ICU stay. Therefore, a robust model should distinguish observed representations from generated ones, adapt its fusion strategy to different modality-availability conditions, and preserve the irregular temporal information encoded in the available clinical streams.

Prompt learning provides a flexible mechanism for adapting model representations to different input conditions through learnable parameters~\cite{gao2021making,khattak2023maple}. Motivated by this property, we propose a multimodal prompt-learning framework for in-hospital mortality prediction using EHR data collected during the first 48 hours of ICU admission. The framework incorporates four complementary prompts. Generative prompts provide surrogate representations for unavailable modalities, missing-signal prompts distinguish observed features from surrogate representations, missing-type prompts characterize modality-availability configurations, and temporal prompts perform condition-specific aggregation over the irregular temporal information. These prompts are jointly integrated into a multimodal Transformer to learn missingness-aware temporal, intramodal, and cross-modal representations from incomplete EHR data. 

We conduct extensive experiments on two large publicly available benchmark datasets, MIMIC-III~\cite{johnson2016mimic} and MIMIC-IV~\cite{johnson2023mimic}, validating the effectiveness and superiority of our proposed method.

The main contributions of this work are three-fold:
\begin{itemize}
    \item We present a novel framework via prompt learning for monitoring critical care patients that is capable of handling missing modalities efficiently with irregular EHR.
    \item We propose four types of prompts to tackle the problem of missing modalities in irregular EHR data. These prompts can generate missing information, learn irregular temporal patterns, and effectively capture both intra- and inter-modality information.
    \item Extensive experiments on large public available benchmark dataset MIMIC-III and MIMIC-IV dataset demonstrate that our proposed approach outperforms the compared methods.
\end{itemize}

%%%%%%%%%%%%%%%%%%%%%  RELATED WORK  %%%%%%%%%%%%%%%%%%%%%%%%%%

\section{Related Work}

\textbf{Multimodal Modeling of Irregular EHRs.}
Electronic health records commonly contain structured multivariate irregular
time series (MISTS) and temporally irregular clinical notes. Earlier studies address irregularity through
time-aware decay, learnable interpolation, set-based representations,
or attention
over continuous-time embeddings~\cite{che2018recurrent,
shukla2019interpolation,horn2020set,shukla2021multi}.
Recent research has increasingly focused on jointly modeling structured
and unstructured EHR modalities. Zhang et al. propose a unified
irregular-EHR framework that models MISTS and clinical-note sequences
separately and integrates them through temporally interleaved
attention~\cite{zhang2023improving}. Liu and Chen further incorporate
domain and memory knowledge into irregular multimodal EHR
representations~\cite{liu2024knowledge}, while CTPD discovers and
aligns cross-modal temporal patterns shared across patients
~\cite{wang2025ctpd}. These approaches improve temporal representation
and multimodal fusion, but primarily focus on extracting information
from the available input streams.

\textbf{Learning with Missing Clinical Modalities.}
Missing modalities are common in clinical data because measurements,
notes, images, and reports are collected under heterogeneous clinical
workflows. Earlier work explored missing-modality learning through
cross-modal translation, modality imagination, and
missing-modality-aware prompting
~\cite{pham2019found,zhao2021missing,lee2023multimodal}. Recent clinical
studies consider more realistic settings: MUSE jointly addresses missing
modalities and labels~\cite{wu2024muse}. Wang et al. address missing
chest radiographs, radiology reports, or structured clinical variables
~\cite{wang2025missing}, while Liang et al. model missing patterns related
to patient conditions or clinical decisions~\cite{liang2025causal}.
More recently, Zhao et al. combine modality reconstruction with
prompt-guided adaptation to handle missing MRI sequences during both
training and inference~\cite{zhao2026dualstage}. Despite these advances,
existing methods primarily emphasize latent compensation, representation
consistency, or robust aggregation. Our work instead couples
prompt-conditioned cross-modal generation with explicit representations
of latent-stream provenances, whole-modality availability, and
irregular temporal progression for paired MISTS and longitudinal
clinical notes.

\textbf{Prompt Learning for Missing Modalities.}
Prompt-based adaptation uses task-specific natural-language templates
in NLP and learnable multimodal prompt tokens in vision-language
models
~\cite{gao2021making,khattak2023maple}. Lee et al. introduces multimodal
prompts for visual recognition with missing inputs
~\cite{lee2023multimodal}. Most closely related to our work, Guo et al.
propose generative, missing-signal, and missing-type prompts for
multimodal sentiment analysis and emotion recognition under missing
modalities~\cite{guo2024multimodal}. However, these methods are designed for general multimodal settings rather than EHR data, and none of them directly addresses missing modalities in irregular EHR with heterogeneous temporal patterns. We adapt this prompt-based missing-modality paradigm to irregular clinical data by introducing temporal prompts and integrating prompt-conditioned reconstruction with time-aware MISTS and clinical-note encoders for ICU outcome prediction.
%%%%%%%%%%%%%%%%%%%%%%%%%%  METHOD  %%%%%%%%%%%%%%%%%%%%%%%%%

\begin{figure*}[htbp!]
  \centering
  \includegraphics[width=\linewidth]{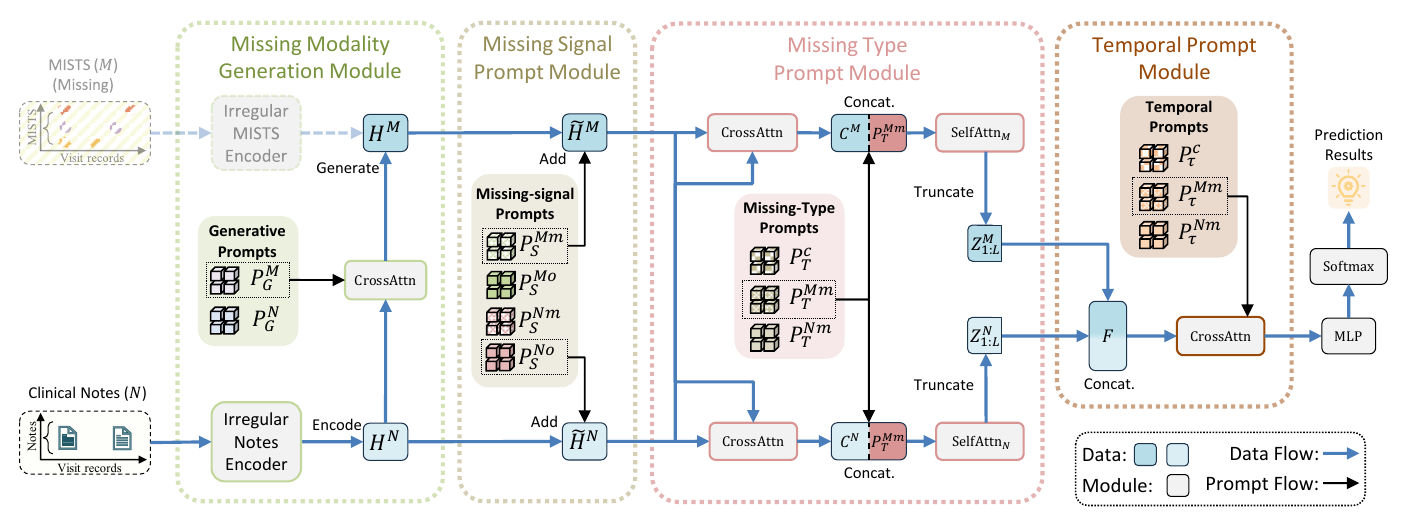}
  \caption{Overview of the proposed prompt-based framework for robust outcome prediction from incomplete irregular multimodal EHRs, based on a Notes-only example. Given a notes-only input condition, irregular notes encoders extract the available representation, and the Missing Modality Generation Module reconstructs the absent MISTS stream. Missing-signal prompts augment each token with a provenance embedding indicating whether its modality stream is observed or prompt-generated, while missing-type prompts inject the sample-level modality-availability condition into both post-fusion streams. The resulting representations are fused and summarized by temporal prompts for patient outcome prediction.}
  \label{fig:framework}
\end{figure*}

\section{Method}

As illustrated in Figure~\ref{fig:framework}, the proposed framework takes multivariate irregularly sampled time series and longitudinal clinical notes as input. Four types of learnable prompts are incorporated into a multimodal Transformer to model irregular temporal information and diverse missing-modality patterns. The resulting multimodal representation is subsequently fed into a prediction head to estimate the probability of in-hospital mortality.

\subsection{Problem Formulation}
We formulate critical care monitoring as a patient-level prediction problem from irregular multimodal EHRs. Each sample from an ICU stay is observed within an early clinical window, that is, the first 48 hours after admission. The sample index is omitted for clarity. The EHR data comprise two heterogeneous modalities: $v\in\{M,N\}$, where $M$ denotes MISTS and $N$ denotes the clinical notes. Let $m\in\{c,Mm,Nm\}$ denote the sample-level modality condition, where $m=c$ indicates that both modalities are observed, $m=Mm$ indicates MISTS is missing, and $m=Nm$ indicates notes is missing. The input can be formulated as
\begin{equation}
    X = \{\{(t_i^M, x^M_i)\}_{i=1}^{T_M}, \{(t_j^N, x^N_j)\}_{j=1}^{T_N}, m\}
\end{equation}
with observation time $t_k^v$, multivariate clinical value $x^v_k$, and number of observed events $T_v$. For a missing modality, the corresponding input sequence is unavailable.
The objective is to learn
    $\hat{y} = f_\theta(X) \in [0,1]$,
the predicted risk of 48-hour in-hospital mortality.

\subsection{Missing Modality Generation Module}
Since MISTS and clinical notes differ in their structures and temporal characteristics, we employ separate modality-specific encoders to obtain their latent representations when they are available: the UTDE encoder~\cite{zhang2023improving} for MISTS and the Clinical-Longformer-based mTAND$_{\mathrm{txt}}$ encoder~\cite{li2022clinical,zhang2023improving} for clinical notes.

\begin{equation}
H^{M}={Enc}_{M}(X^{M},T^{M})
\end{equation}
\begin{equation}
H^{N}={Enc}_{N}(X^{N},T^{N})
\end{equation}

where $X^{M}$, $T^{M}$ denote the numerical observations, timestamps of MISTS, respectively, while $X^{N}$ and $T^{N}$ denote the note embeddings and their timestamps.

When one modality is missing, we employ the Missing Modality Generation Module to construct a surrogate representation for the unavailable modality. As depicted in Figure~\ref{fig:framework}, when clinical notes are observed and MISTS are missing ($m=Mm$), $H^{N}$ is obtained directly from the irregular notes encoder, while $H^{M}$ is generated from $H^{N}$ using the generative prompt $P_G^{M}$ through cross-attention~\cite{vaswani2017attention}:

\begin{equation}
    H^{M}
    =
    \operatorname{CrossAttn}
    \left(P_G^{M},H^{N},H^{N}\right).
\end{equation}

Here, $P_G^{M}\in\mathbb{R}^{L\times d}$ is a learnable generative prompt, where $L$ denotes the number of positions on the unified temporal grid and $d$ denotes the latent dimension shared by the two modalities. $P_G^{M}$ serves as the query, while the observed clinical-note representation $H^{N}$ serves as both the keys and values. The resulting $H^{M}$ is a time-aligned surrogate representation for the missing MISTS modality.

\subsection{Missing Signal Prompt Module}

Although the generated representation $H^{M}$ restores the missing MISTS stream, it should not be treated in the same way as the directly observed clinical-note representation $H^{N}$. We therefore introduce the Missing Signal Prompt Module to indicate whether each modality representation is observed or generated.

The module annotates each stream with its provenance using four learnable prompts for both modalities: $P^{Mm}_{S}, P^{Mo}_{S}, P^{Nm}_{S}, P^{No}_{S} \in \mathbb{R}^{L\times d}$, indicating that the modality is missing ($Mm, Nm$) or observed ($Mo, No$). Therefore, the corresponding prompts are added as follows:

\begin{align}
    \widetilde{H}^{M}
    &= H^{M}+P_{S}^{Mm},\\
    \widetilde{H}^{N}
    &= H^{N}+P_{S}^{No}.
\end{align}

Here, by adding $P_{S}^{Mm}$ and $P_{S}^{No}$ to the generated MISTS representation and observed clinical-note representation, respectively, each time-aligned token now carries a position-specific marker of its source, so that the downstream fusion is able to weigh the observed and synthesized features asymmetrically rather than blending them uniformly.

\subsection{Missing Type Prompt Module}
Although the pair of provenance prompts implicitly identifies the modality condition, this information remains distributed across the two streams. We therefore introduce a shared missing-type prompt that provides both post-fusion branches with an explicit global availability condition.

Specifically, the Missing Type Prompt Module provides an explicit global condition through three learnable missing-type prompts: $P^{c}_{T}, P^{Mm}_{T}, \, P^{Nm}_{T} \in \mathbb{R}^{L_p \times d}$, where $L_p$ is the number of type tokens. The same missing-type prompt is used for both modality streams to provide a global condition.

The two modality representations first exchange information through bidirectional cross-attention:

\begin{align}
    C^{M}
    &=
    \operatorname{CrossAttn}
    \left(\widetilde{H}^{M},
          \widetilde{H}^{N},
          \widetilde{H}^{N}\right),\\
    C^{N}
    &=
    \operatorname{CrossAttn}
    \left(\widetilde{H}^{N},
          \widetilde{H}^{M},
          \widetilde{H}^{M}\right).
\end{align}

% We then append the selected type prompt to each cross-attended sequence and refine the resulting sequence with modality-specific self-attention:
When modality $M$ is missing, $P_T^{Mm}$ is selected and appended to both cross-attended representations:

\begin{align}
    Z^{M}
    &=
    \operatorname{SelfAttn}_{M}
    \left([C^{M};P_T^{Mm}]\right),\\
    Z^{N}
    &=
    \operatorname{SelfAttn}_{N}
    \left([C^{N};P_T^{Mm}]\right).
\end{align}

% Here $\mathrm{SelfAttn}_v$ denotes a modality-specific Transformer stack operating on a sequence of length $L+L_p$. The two stacks use the same architecture but have separate parameters.
Here, $[\cdot\,;\cdot]$ denotes sequence-wise concatenation and $Z^{M},Z^{N}\in\mathbb{R}^{(L+L_p)\times d}$. The two self-attention stacks share the same architecture but have separate parameters. By using $P_T^{Mm}$ in both streams, the fusion module is explicitly informed that the MISTS representation is generated while the clinical-note representation is observed.

\subsection{Temporal Prompt Module}
The final prediction still requires a patient-level summary over the irregular clinical timeline. A fixed pooling operation cannot adaptively identify the temporal positions that are most relevant under different modality-availability conditions. We therefore introduce the Temporal Prompt Module for condition-specific temporal aggregation.

Because $Z^{M}$ and $Z^{N}$ contain both temporal tokens and the appended missing-type prompt tokens, we retain their first $L$ temporal tokens and concatenate them along the feature dimension:

\begin{equation}
    F
    =
    \operatorname{Concat}_{\mathrm{feat}}
    \left(Z^M_{1:L},Z^N_{1:L}\right)
    \in\mathbb{R}^{L\times 2d}.
\end{equation}

Here, $Z^M_{1:L}$ and $Z^N_{1:L}$ denote the first $L$ time-aligned modality tokens in $Z^M$ and $Z^N$. The remaining $L_p$ tokens correspond to the appended missing-type prompts and are excluded from temporal aggregation. $\operatorname{Concat}_{\mathrm{feat}}$ denotes concatenation along the feature dimension.

To adapt temporal aggregation to different modality-availability conditions, we maintain three learnable temporal prompts: $ P^c_\tau,\, P^{Mm}_{\tau},\, P^{Nm}_{\tau}\in \mathbb{R}^{1\times 2d}$, corresponding to three modality missing condition. When modality $M$ is missing, $P_\tau^{Mm}$ is selected:

\begin{equation}
    s_{Mm}
    =
    \operatorname{CrossAttn}
    \left(P_\tau^{Mm},F,F\right).
\end{equation}
Here, $P_\tau^{Mm}$ serves as the query, while $F$ provides the keys and values. The resulting representation $s_{Mm}\in\mathbb{R}^{1\times 2d}$ summarizes the temporal information most relevant to the Notes-only condition.

Finally, the summarized representation is processed by a residual two-layer MLP and a linear classifier:
\begin{align}
r_{Mm}
&=
s_{Mm}
+
\operatorname{MLP}(s_{Mm}),\\
\widehat{y}
&=
\left[
\operatorname{Softmax}
\left(W_o r_{Mm}+b_o\right)
\right]_{\mathrm{mortality}}
\end{align}
Here, ${r}_{Mm}\in\mathbb{R}^{2d}$ denotes the resulting patient-level representation after residual connection. $W_o\in\mathbb{R}^{2\times 2d}$ and ${b}_o\in\mathbb{R}^{2}$ are the learnable weight matrix and bias vector of the binary classifier, respectively. $\widehat{y}\in[0,1]$ denotes the predicted probability of in-hospital mortality.

\section{Experiment}
\subsection{Datasets}
We evaluate our method on multimodal ICU cohorts from MIMIC-III~\cite{johnson2016mimic} and MIMIC-IV v3.1.\footnotemark[1]\footnotetext[1]{\url{https://physionet.org/content/mimiciv/3.1/}} For MIMIC-III, we follow the processed irregular-EHR benchmark cohort of Zhang et al.~\cite{zhang2023improving}; for MIMIC-IV, clinical notes are taken from MIMIC-IV-Note.\footnotemark[2]\footnotetext[2]{\url{https://physionet.org/content/mimic-iv-note/2.2/}} The task is 48-hour in-hospital mortality prediction (48-IHM), where each sample uses the first 48 hours after ICU admission to predict hospital mortality. It is a binary classification
problem with label imbalance with death to discharge ratio
of approximately 1:7. ICU stays shorter than 48 hours are excluded.

Each sample contains a structured MISTS modality and a clinical-note modality. MISTS includes 17 benchmark clinical variables with irregular timestamps and observation masks; we also build a one-hour value-and-mask representation and normalize continuous variables using training-set statistics. Clinical notes are aligned by ICU admission time, and at most the latest five notes within the 48-hour window are tokenized with Clinical-Longformer and truncated or padded to 1024 tokens per note. 

\subsection{Evaluation Metrics}
Following prior work~\cite{zhang2023improving}, we evaluate 48-IHM as
an imbalanced binary prediction task in which identifying mortality cases
is clinically important. We report F1 to characterize classification
performance at a fixed decision threshold, AUPRC to evaluate positive-case
retrieval under label imbalance, and AUROC to provide a complementary
measure of overall discrimination. Because the dataset is class-imbalanced and mortality is the minority class, we use AUPRC as the primary metric for model selection, as it better reflects the model's ability to identify mortality cases.

\subsection{Baselines}
To ensure a comprehensive comparison, we evaluate our method against three groups of baselines: MISTS-only approaches, Notes-only approaches, and Full-modality approaches. For MIMIC-III, we follow the published irregular-EHR baseline results ~\cite{zhang2023improving}; for MIMIC-IV, we evaluate the listed baselines under our data split.

For the \emph{MISTS-only} setting on MIMIC-III, we follow the benchmark comparison with Imputation~\cite{lipton2016directly}, IP-Net~\cite{shukla2019interpolation}, mTAND~\cite{shukla2021multi}, GRU-D~\cite{che2018recurrent}, RAINDROP~\cite{zhang2022graphguided}, and UTDE~\cite{zhang2023improving}. On MIMIC-IV, we evaluate the regular time-series backbones IP-Net, GRU-D, SeFT~\cite{horn2020set}, DGM$^2$-O~\cite{wu2021dynamic}, mTAND, RAINDROP, MTGNN~\cite{wu2020connecting}, and UTDE.

For the \emph{Notes-only} setting on both datasets, we compare with the note-based models actually used in our experiments: Flat~\cite{deznabi2021predicting}, HierTrans~\cite{pappagari2019hierarchical}, T-LSTM~\cite{baytas2017patient}, FT-LSTM~\cite{zhang2020time}, GRU-D~\cite{che2018recurrent}, and mTAND-txt~\cite{zhang2023improving}.

For the \emph{Full-modality} setting on MIMIC-IV, we compare with representative multimodal EHR and feature-fusion baselines, including MMIMIC~\cite{zhang2023improving}, CTPD~\cite{wang2025ctpd}, MedFuse~\cite{hayat2022medfuse}, DAFT~\cite{polsterl2021combining}, MMTM~\cite{joze2020mmtm}, and DrFuse~\cite{yao2024drfuse}. Under each missing-modality setting, our model receives only the available modality and reconstructs the missing stream through prompt-based cross-modal generation.

\subsection{Implementation Details}
\paragraph{Architecture and Training Details.}

We use a shared embedding dimension of $d=128$ and a unified timeline of length $T=48$. 
Clinical notes are pre-tokenized by Clinical-Longformer with a maximum token length of 1024, and are temporally aligned to the unified timeline using mTAND-txt. For the MISTS stream, we adopt UTDE, which incorporates a learnable gate based on mTAND-ts.
%CA details
The text encoder is initialized from a pretrained Clinical-Longformer checkpoint, while the remaining task-specific modules are trained from scratch. Each run is trained up to a fixed maximum number of epochs, and model selection uses validation AUPRC under the corresponding missing-modality scenario.

\paragraph{Prompt initialization.}
Generative prompt queries are initialized from the training-set global means of the projected target-modality embeddings. The modality-signal and missing-type prompts are initialized to zero, whereas the missing-mode-specific temporal prompts are independently initialized from a standard normal distribution. All prompt parameters are jointly optimized with the remaining task-specific modules.

\paragraph{Training-Time Modality Dropout.}
To expose the model to incomplete inputs during training, we apply modality-level dropout to the paired multimodal training samples. For each complete training batch, we sample a Bernoulli variable with probability $\eta$. If dropout is not activated, both MISTS and clinical notes are kept and the sample is trained under the complete case $m=\mathrm{c}$. If dropout is activated, one modality is removed at the input level according to the target missing scenario. 

\subsection{Main Results}
We first examine our method's robustness when either modality is unavailable, followed by assessing whether the proposed framework retains competitive performance when both modalities are observed.

\begin{table*}[htbp!]
\caption{Main results on MIMIC-III under single-modality settings. Best results are shown in bold and second-best results are underlined.}
  \centering
  {\small
  \setlength{\tabcolsep}{1pt}
  \begin{tabular}{@{}l*{14}{c}@{}}
    \toprule
    & \multicolumn{7}{c}{\textit{MISTS-only setting}} & \multicolumn{7}{c}{\textit{Notes-only setting}} \\
    \cmidrule(lr){2-8} \cmidrule(lr){9-15}
    Metric & Imputation & IP-Net & mTAND & GRU-D & RAINDROP & UTDE & Ours & Flat & HierTrans & T-LSTM & FT-LSTM & GRU-D & mTAND-txt & Ours \\
    \midrule
    F1 & 0.3973 & 0.3722 & 0.4387 & 0.4282 & 0.3946 & \underline{0.4526} & \textbf{0.4961} & 0.3978 & 0.4876 & 0.5032 & 0.4851 & 0.5101 & \underline{0.5257} & \textbf{0.5284} \\
    AUPRC & 0.4436 & 0.3936 & 0.4754 & 0.4590 & 0.3623 & \underline{0.4964} & \textbf{0.5028} & 0.5169 & 0.5298 & 0.5257 & 0.5439 & 0.5434 & \underline{0.5605} & \textbf{0.5853} \\
    \bottomrule
  \end{tabular}%
  }
  
  \label{tab:main_mimic3}
\end{table*}

\begin{table}[htbp!]
\caption{Main results on MIMIC-IV under single-modality settings. Best results within each setting and metric are shown in bold and second-best results are underlined.}
\centering
\small
\setlength{\tabcolsep}{3pt}

% ---- MISTS-only subtable ----
\begin{tabular}{l|*{9}{c}}
\toprule
\multicolumn{10}{c}{\textit{MISTS-only setting}} \\
\midrule
Metric & GRU-D & IP-Net & SeFT & MTGNN & DGM$^2$-O & mTAND & RAINDROP & UTDE & Ours \\
\midrule
AUROC & 0.8293 & 0.8580 & 0.6536 & 0.8314 & 0.8394 & 0.8540 & 0.8323 & \textbf{0.8620} & \underline{0.8586} \\
AUPRC & 0.4541 & \underline{0.5154} & 0.2369 & 0.4826 & 0.4832 & 0.5025 & 0.4447 & 0.5099 & \textbf{0.5202} \\
F1    & 0.3114 & 0.4044 & 0.0672 & 0.4191 & 0.4190 & 0.4168 & 0.3219 & \textbf{0.4856} & \underline{0.4585} \\
\bottomrule
\end{tabular}

\medskip

% ---- Notes-only subtable ----
\begin{tabular}{l|*{7}{c}}
\toprule
\multicolumn{8}{c}{\textit{Notes-only setting}} \\
\midrule
Metric & T-LSTM & GRU-D & HierTrans & FT-LSTM & Flat & mTAND-txt & Ours \\
\midrule
AUROC & 0.6748 & 0.8120 & \underline{0.8148} & 0.8109 & 0.8112 & 0.8099 & \textbf{0.8181} \\
AUPRC & 0.2677 & 0.4329 & 0.4380 & \underline{0.4384} & 0.4245 & 0.4170 & \textbf{0.4509} \\
F1    & 0.1394 & \underline{0.4121} & 0.3798 & 0.3937 & 0.4081 & 0.2229 & \textbf{0.4643} \\
\bottomrule
\end{tabular}

\label{tab:main_mimic4}
\end{table}
\paragraph{Missing-modality Comparison}
To examine whether the proposed framework can recover useful information beyond that captured by unimodal models, Tables~\ref{tab:main_mimic3} and~\ref{tab:main_mimic4} compare our method with strong MISTS-only and Notes-only baselines on two independent cohorts. Our method achieves the highest AUPRC in all four settings. Since mortality cases are underrepresented and AUPRC is our primary metric, this consistent result provides strong evidence that our prompt-based framework improves the identification and ranking of high-risk patients under missing modalities. Moreover, the gains are more consistent across metrics in the Notes-only setting than in the MISTS-only setting. This asymmetry suggests that reconstructing MISTS from clinical notes is generally more stable than reconstructing clinical notes from MISTS.

\begin{table}[htbp!]
\caption{Full-modality results on MIMIC-IV, where both MISTS and clinical notes are available at test time. Best results are shown in bold and second-best results are underlined.}
\centering
\small
\setlength{\tabcolsep}{3pt}
\begin{tabular}{l|ccccccc}
\toprule
Metric & MMTM & DAFT & MedFuse & DrFuse & CTPD & MMIMIC & Ours \\
\midrule
AUROC & 0.8776 & 0.8488 & 0.8523 & 0.8588 & 0.8694 & \textbf{0.8826} & \underline{0.8785} \\
AUPRC & \underline{0.5761} & 0.4910 & 0.5041 & 0.5141 & 0.5384 & \textbf{0.5767} & 0.5587 \\
F1    & 0.4823 & 0.2541 & 0.3614 & 0.4698 & 0.4060 & \underline{0.4839} & \textbf{0.5285} \\
\bottomrule
\end{tabular}
  \label{tab:main_mimic4_full}
\end{table}

\paragraph{Full-modality Comparison.}
We further evaluate whether our method maintains its performance when both modalities are available. As shown in Table~\ref{tab:main_mimic4_full}, our method achieves the best F1 and the second-best AUROC. Compared with MMIMIC, which serves as our backbone, it substantially improves F1, with only modest decreases in AUROC and AUPRC. These results indicate that the proposed method largely preserves the backbone's predictive capability under complete observations in addition to handling missing modalities effectively.

\subsection{Ablation Study}
\label{sec:ablation}
To better understand the contribution of each design choice, we conduct ablation studies from two perspectives: contribution of different prompts and training-time modality dropout rate.

\begin{table}[htbp!]
\caption{Prompt ablation under the Notes-only setting on MIMIC-III, including individual, three-prompt, and cumulative configurations. A check mark indicates that the corresponding prompt is enabled. Best and second-best results are shown in bold and underlined, respectively.}
  \centering
  {\small
  \setlength{\tabcolsep}{7pt}
  \begin{tabular}{cccccc}
    \toprule
    \multicolumn{4}{c}{Prompt Configuration} &
    \multicolumn{2}{c}{Performance} \\
    \cmidrule(lr){1-4} \cmidrule(lr){5-6}
    $P_G$ & $P_S$ & $P_T$ & $P_\tau$ & AUPRC & AUROC \\
    \midrule
     &  &  &  & 0.5383 & 0.8759 \\
    $\checkmark$ &  &  &  & 0.5566 & 0.8841 \\
     & $\checkmark$ &  &  & 0.5572 & 0.8839 \\
     &  & $\checkmark$ &  & 0.5527 & 0.8822 \\
     &  &  & $\checkmark$ & 0.5412 & 0.8782 \\
    \midrule
    $\checkmark$ & $\checkmark$ &  &
        & 0.5695 & 0.8864 \\
    $\checkmark$ & $\checkmark$ & $\checkmark$ &
        & \underline{0.5768} & \textbf{0.8917} \\
    $\checkmark$ & $\checkmark$ & $\checkmark$ & $\checkmark$
        & \textbf{0.5853} & \underline{0.8904} \\
    \midrule
     & $\checkmark$ & $\checkmark$ & $\checkmark$
        & 0.5597 & 0.8874 \\
    $\checkmark$ &  & $\checkmark$ & $\checkmark$
        & 0.5700 & 0.8872 \\
    $\checkmark$ & $\checkmark$ &  & $\checkmark$
        & 0.5687 & 0.8875 \\
    \bottomrule
  \end{tabular}
  }

  \label{tab:prompt_ablation}
\end{table}

\paragraph{Prompt Ablation.}
Table~\ref{tab:prompt_ablation} demonstrates the effectiveness and complementarity of the four prompts. Enabling any single prompt yields modest improvements in both AUPRC and AUROC, indicating that each provides useful task-specific information. The cumulative configurations show a clear progression: $P_G$ reconstructs the missing modality, $P_S$ identifies the provenance of generated and observed features, $P_T$ adapts cross-modal interaction to the modality-availability pattern, and $P_\tau$ performs condition-specific temporal aggregation. As these functions are progressively integrated, AUPRC increases steadily and reaches its highest value of 0.5853 with all prompts enabled. Consistently, removing any prompt from the complete configuration reduces AUPRC to varying degrees, supporting that the four prompts make non-redundant contributions and work together to improve missing-modality prediction.

\begin{figure}[htbp!]
  \centering
  \includegraphics[width=0.7\linewidth]{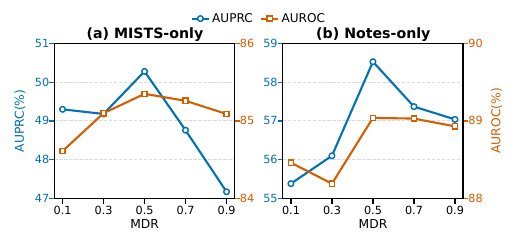}
  \caption{Modality dropout rate ablation on MIMIC-III. Each panel reports AUPRC on the left y-axis and AUROC on the right y-axis for $\eta\in\{0.1,0.3,0.5,0.7,0.9\}$.}
  \label{fig:mdr_ablation}
\end{figure}

\paragraph{Dropout Rate.}
We vary the training-time modality dropout rate $\eta$ to examine how exposure to simulated missing inputs affects robustness. As shown in Figure~\ref{fig:mdr_ablation}, both settings show a non-monotonic trend. When the dropout rate is too low, the model is not sufficiently exposed to incomplete cases during training; when it is too high, effective observed supervision is reduced and the generated representations become less reliable. These observations suggest that a moderate-to-high dropout rate forces the model to learn stronger missing-modality compensation and more robust cross-modal interaction under incomplete observations.

\paragraph{Representation Visualization.}
\begin{figure}[htbp!]
  \centering
  \includegraphics[width=0.6\linewidth]{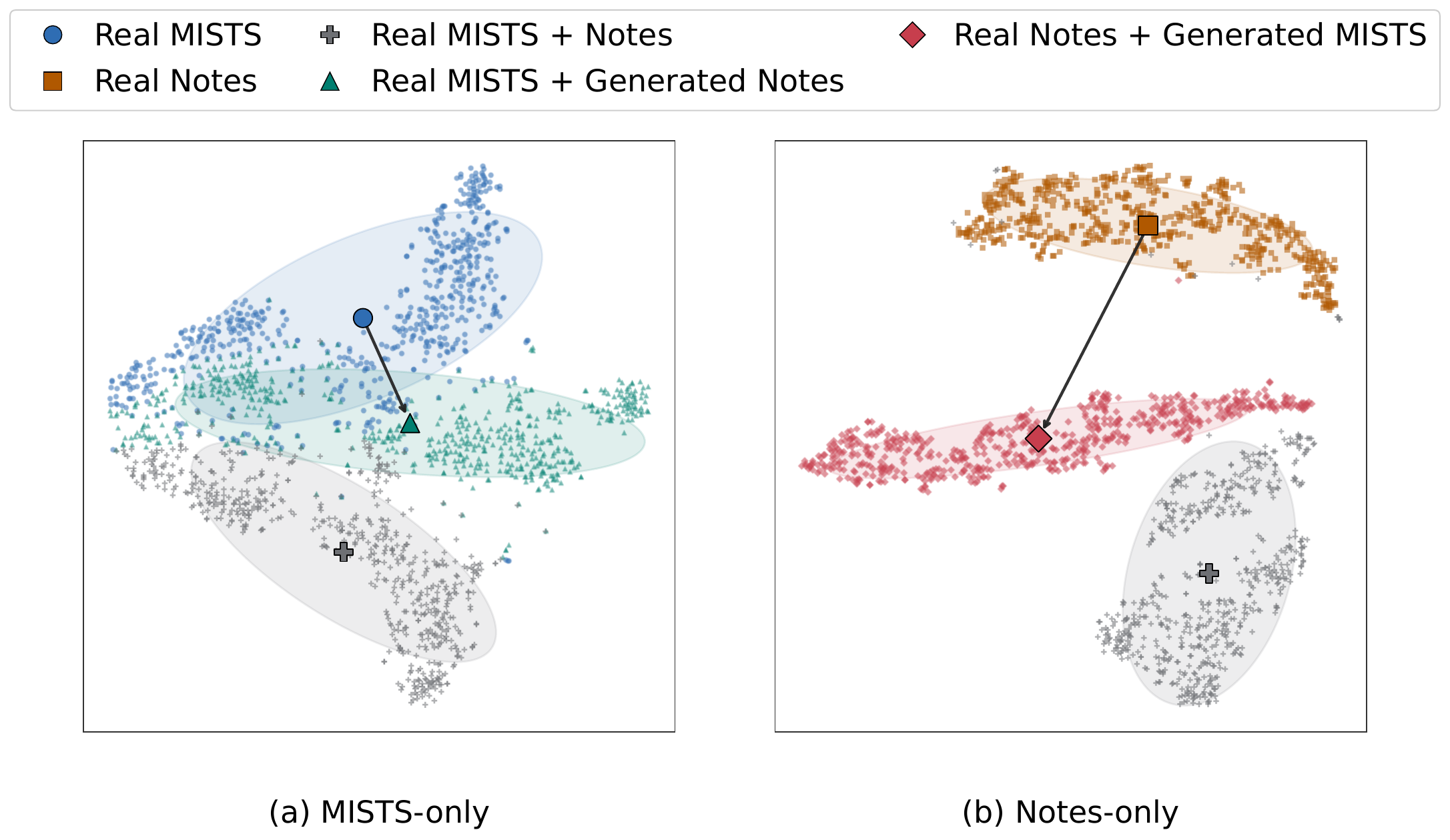}
  \caption{t-SNE visualization of classifier-level representations on MIMIC-III. Panel (a) and (b) show the MISTS-only and Notes-only settings respectively. Each panel compares the observed single-modality representation, its prompt-completed representation, and the fully observed MISTS+Notes reference.}
  \label{fig:completion_tsne}
\end{figure}

To examine whether prompt completion recovers meaningful cross-modal structure, we visualize the learned representations on MIMIC-III. As shown in Figure~\ref{fig:completion_tsne}, the completed representations shift from the single-modality region toward the Full-modality region while retaining the structure of the observed modality. This controlled shift indicates that prompt completion supplements rather than overwrites the available evidence, reducing the representation gap between incomplete and complete EHR inputs.

\begin{figure}[htbp!]
  \centering
  \includegraphics[width=0.4\linewidth]{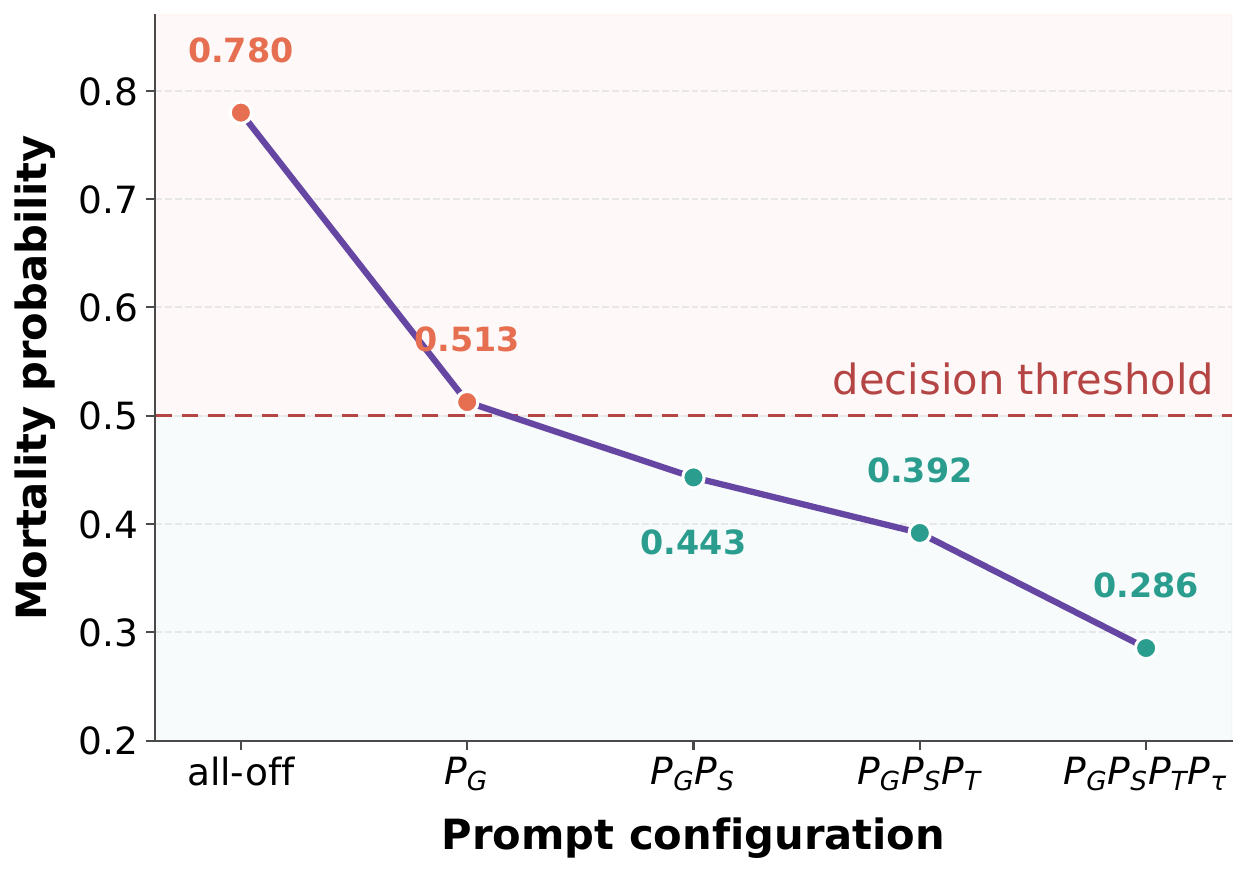}
  \caption{Evolution of the predicted mortality risk for a representative patient in the MIMIC-III Notes-only setting as the prompt components are cumulatively activated. \textit{All-off} indicates that no prompts are used. Predictions above the decision threshold are classified as mortality, whereas those below the threshold are classified as survival.}
  \label{fig:case_study}
\end{figure}

\begin{figure}[htbp]
  \centering
  \includegraphics[width=0.7\linewidth]{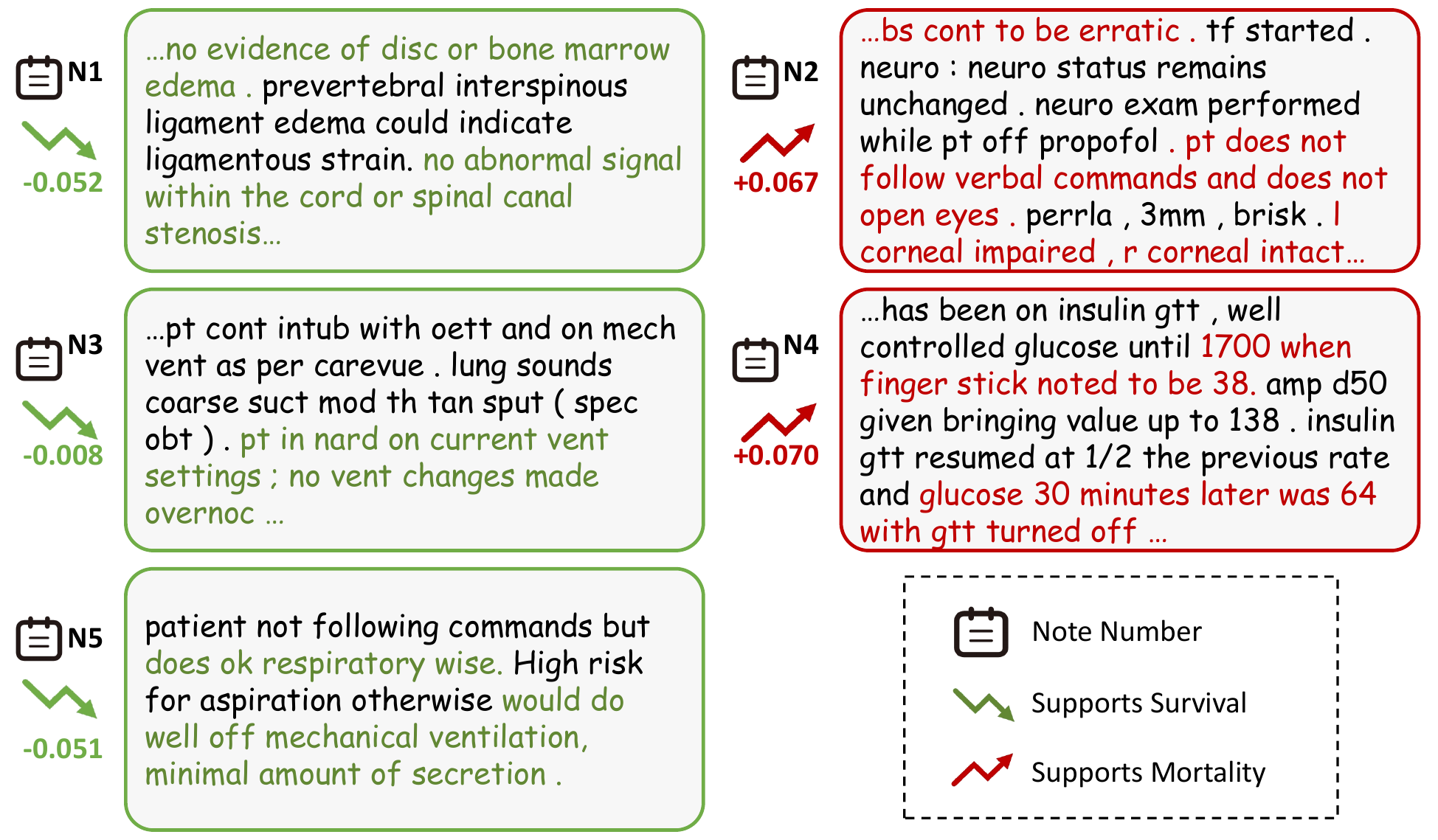}
  \caption{Changes in predicted mortality risk after removing each note for the same patient. Negative scores shown in green indicate evidence supporting survival, whereas positive scores shown in red indicate evidence supporting mortality.}
  \label{fig:case_study_notes}
\end{figure}

\subsection{Case Study}
To examine how the prompts affect individual predictions and how the model integrates evidence across clinical notes, we conduct a case study on a representative patient. Figure~\ref{fig:case_study} illustrates how the prompts correct a survival case under the MIMIC-III Notes-only setting. Without prompts, the model incorrectly predicts mortality. $P_G$ produces the largest risk reduction, showing that the reconstructed MISTS stream supplies the main complementary evidence. $P_S$ then moves the prediction below the decision threshold, while $P_T$ and $P_\tau$ maintain the correct decision. This progression indicates that reconstruction drives the correction, whereas the remaining prompts refine how generated and observed evidence is interpreted.

Figure~\ref{fig:case_study_notes} further shows that the prediction is based on evidence distributed across multiple notes. The impaired responsiveness in N2 and severe hypoglycemia in N4 support mortality, whereas N1 and N5 contain evidence favoring survival. The model therefore reaches the correct prediction by balancing conflicting clinical observations rather than relying on a single note.

\section{Conclusion}
In this paper, we present a unified multimodal prompt-learning framework for robust mortality prediction from irregular and incomplete EHRs. We proposed generative prompts, missing-signal prompts, missing-type prompts, and temporal prompts to capture information under missing-modality conditions. Ablation study and case study further confirm that the four prompts play complementary roles in aligning incomplete inputs with the full-modality decision space. Future work can evaluate the framework on broader clinical prediction tasks and datasets, and extend it to settings involving more modalities and complex real-world missingness patterns.

\bibliographystyle{unsrt}  
\bibliography{references}

\end{document}